\documentclass[letterpaper, 10 pt, conference]{ieeeconf}  

\IEEEoverridecommandlockouts                              

\usepackage{booktabs}
\usepackage{adjustbox}
\usepackage{xcolor}
\usepackage{graphicx}
\usepackage{amsmath}
\usepackage{float}
\usepackage{caption}
\usepackage{lipsum}
\usepackage{array} 
\usepackage{subcaption}
\usepackage{balance}
\usepackage[bookmarks=true]{hyperref}

\begin{document}
\title{\LARGE \bf
Humanoids in Hospitals: A Technical Study of Humanoid Robot Surrogates for Dexterous Medical Interventions
}

\author{Soofiyan Atar$^{1}$, Xiao Liang$^{*,1}$, Calvin Joyce$^{*,1}$, Florian Richter$^{1}$, Wood Ricardo$^{2}$, \\ Charles Goldberg$^{2}$, Preetham Suresh$^{2}$, Michael Yip$^{1}$, \IEEEmembership{Senior Member, IEEE}
\thanks{$^*$ \textit{These authors contributed equally to this work}}
\thanks{$^1$Electrical and Computer Engineering Department, University of California San Diego, La Jolla, CA 92093 USA. {\tt\footnotesize\{satar, x5liang, cajoyce,  frichter, m1yip\}@ucsd.edu}}%
\thanks{$^2$ University of California San Diego, La Jolla, CA 92093 USA. {\tt\footnotesize\{r2wood, cggoldberg, pjsuresh, \}@health.ucsd.edu}}%
}

\let\oldtwocolumn\twocolumn
\renewcommand\twocolumn[1][]{%
    \oldtwocolumn[{#1}{
    \begin{center}
           \vspace{-4mm}
           \includegraphics[width=0.99\textwidth]{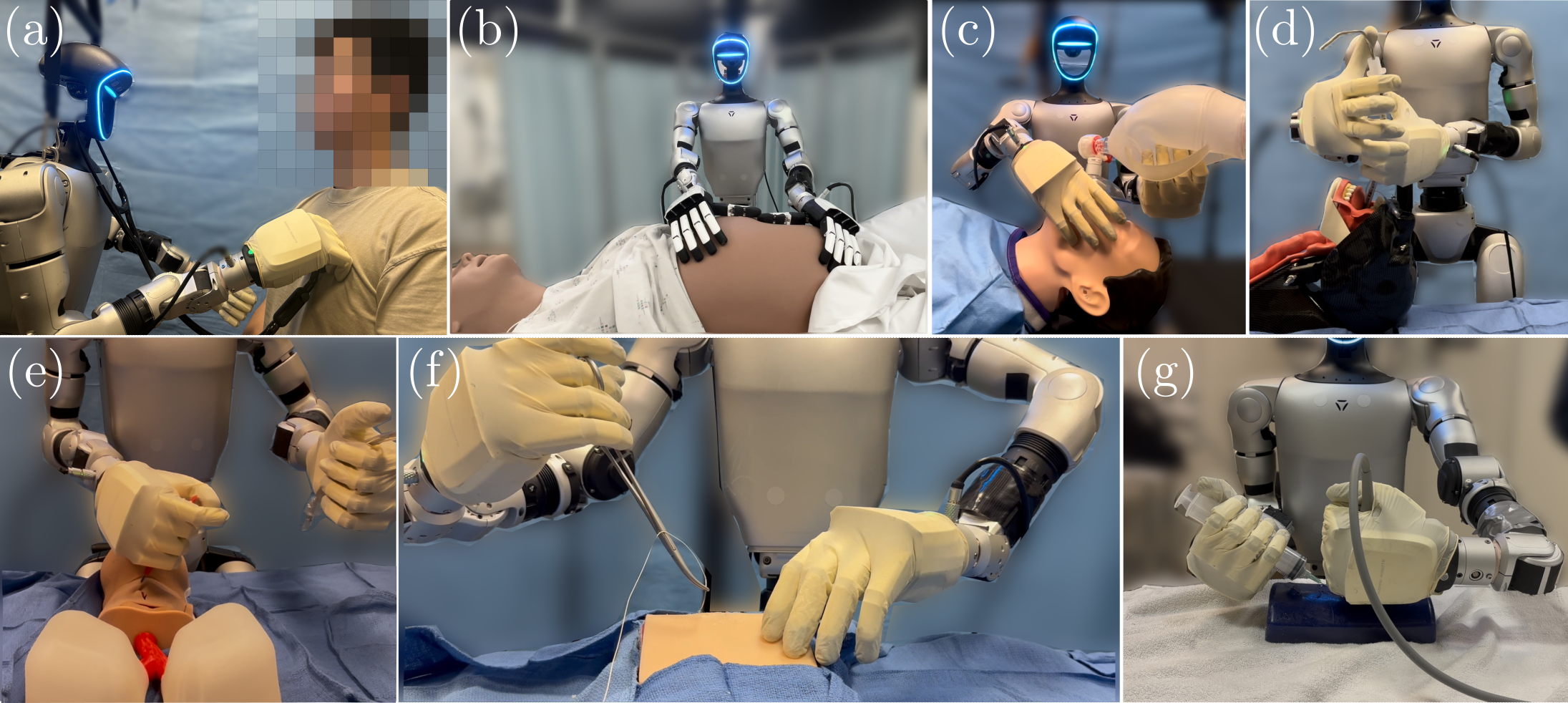}
           \captionof{figure}{\textbf{Teleoperated humanoid robot in diverse medical scenarios.} The following were performed with the presented teleoperation system: (a). auscultation, (b). Leopold maneuvers (abdominal palpation), (c) bag valve mask ventilation, (d). endotracheal intubation, (e). tracheostomy, (f). suturing throw, and (g). ultrasound-guided injection.}
           \label{fig:fig1}
        \end{center}
    }]
}


\maketitle
\begin{abstract}
The increasing demand for healthcare workers, driven by aging populations and labor shortages, presents a significant challenge for hospitals.
Humanoid robots have the potential to alleviate these pressures by leveraging their human-like dexterity and adaptability to assist in medical procedures.
This work conducted an exploratory study on the feasibility of humanoid robots performing direct clinical tasks through teleoperation.
A bimanual teleoperation system was developed for the Unitree G1 Humanoid Robot, integrating high-fidelity pose tracking, custom grasping configurations, and an impedance controller to safely and precisely manipulate medical tools.
The system is evaluated in seven diverse medical procedures, including physical examinations, emergency interventions, and precision needle tasks.
Our results demonstrate that humanoid robots can successfully replicate critical aspects of human medical assessments and interventions, with promising quantitative performance in ventilation and ultrasound-guided tasks.
However, challenges remain, including limitations in force output for procedures requiring high strength and sensor sensitivity issues affecting clinical accuracy.
This study highlights the potential and current limitations of humanoid robots in hospital settings and lays the groundwork for future research on robotic healthcare integration. \href{https://surgie-humanoid.github.io}{Website link}
\end{abstract}


\section{INTRODUCTION}

Societal pressures such as aging populations \cite{ismail2021impact} and labor shortages \cite{dzakula2022shortage} have bolstered the need for human-like robotic systems to assist or augment human workers~\cite{doncieux2022human, gigliotta2024hugo}.
Coupled with recent advances in artificial intelligence \cite{doncieux2022human}, bipedal control \cite{mikolajczyk2022bipedal}, actuator development \cite{kong2009actuator}, and sensor technology \cite{zhang2017human}, there is an explosion of interest in humanoid systems that can operate directly in environments structured for humans~\cite{meta2025humanoid, apptronik2025funding}.
Assembly and production tasks are a primary target for humanoids since their human-like form allows them to maneuver in industrial workcells and use the same tools or interfaces as human workers \cite{ubtech_walkers}.

Beyond structured environments, humanoids are expected to play key roles in maintenance and hazardous duties \cite{hazardous}, security and facility services \cite{security}, space exploration \cite{space}, and disaster response \cite{disaster}, where humanistic adaptability is crucial. Furthermore, their ability to mimic human expressions enhances interactions in hospitality \cite{app142411823}, customer service \cite{8734250}, retail \cite{song2022role}, education \cite{ekstrom2022dual}, and human-robot cognition research \cite{cross2020social}, making them valuable both in the technical and social domains.

Unlike humanoids, robotic systems deployed in healthcare are morphologically specialized to serve their narrow roles. Examples include robots for laparoscopic surgery \cite{ngu2017da}, orthopedics \cite{roche2021mako}, and bronchoscopy \cite{lu2021review}.
However, a significant gap remains between the capabilities of these robotic systems and the growing labor demands in hospitals today.
Today, hospitals face mounting labor shortages, with an increasing demand for healthcare workers driven by aging populations \cite{jones2024healthcare}, rising patient volumes \cite{payet2020influence}, and workforce burnout \cite{OSG2022burnout}.
This has led to longer wait times in emergency departments \cite{goodacre2005who}, critical nursing shortages that strain patient care \cite{tamata2023systematic}, and a lack of physicians in rural areas, where some communities have limited or no access to specialized medical services \cite{ahrq2021nhqdr}.
We believe that the flexibility provided by modern humanoid systems is key to alleviating the growing demand for labor in hospitals.

Previous work on humanoid robots in healthcare has focused mainly on nonclinical tasks rather than directly automating clinical procedures. For example, Moxi \cite{diligentrobots_moxi} was designed to help transport items in hospitals, and \cite{joseph2018review} \cite{9850577} reviews systems that help with tasks such as flipping a person in a medical bed \cite{5980213}, moving patients \cite{5651735}, and setting or cleaning \cite{5569012}. In contrast, humanoid robots are increasingly being used to address the challenges posed by an aging population; \cite{maa2022humanoid} demonstrated their use in teleoperated reminiscence group therapy for older adults with dementia, capitalizing on their ability to read facial expressions and provide personalized care. Unlike existing robots, the teleoperated humanoid robot explored in this work performs medical tasks in a simulated setting, including physical examinations and emergency interventions.


This work explores the feasibility of humanoid robots performing medical procedures in hospital settings.
A teleoperation system capable of grasping and manipulating various medical tools is developed, which allows humanoid robots to perform various clinical tasks. Several enabling methods are required to ensure that various objects can be manipulated and procedures can be performed. An exploratory study evaluates the usability of humanoid robots in performing routine physical examinations, emergency interventions, and precision needle tasks, providing a detailed discussion of their limitations and potential role in modern healthcare.

\section{METHODS}
\begin{figure}[t]
    \centering
    \includegraphics[width=\linewidth]{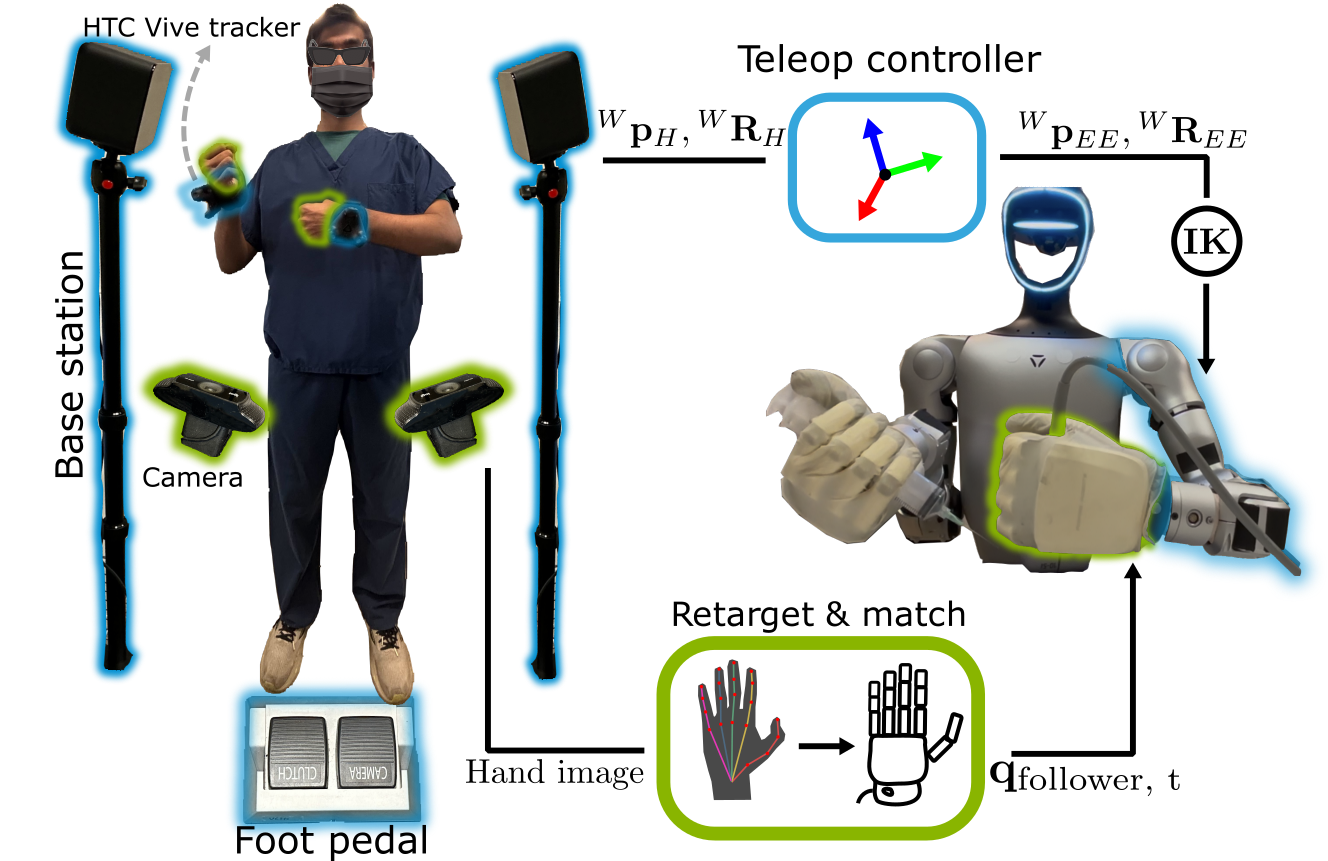} 
    \caption{\textbf{A visualization of the proposed humanoid teleoperation interface.} An operator uses HTC Vive trackers along with a foot pedal to control the robot hands' transformation (shown in blue contours). The hands' configuration is changed through re-targeting human hand pose from webcam images to robot hand configurations (shown in green contours).}
    \label{fig:teleop_pipeline}
    \vspace{-2.em}
\end{figure}

\begin{figure}[t]
    \centering
    \includegraphics[width=\linewidth]{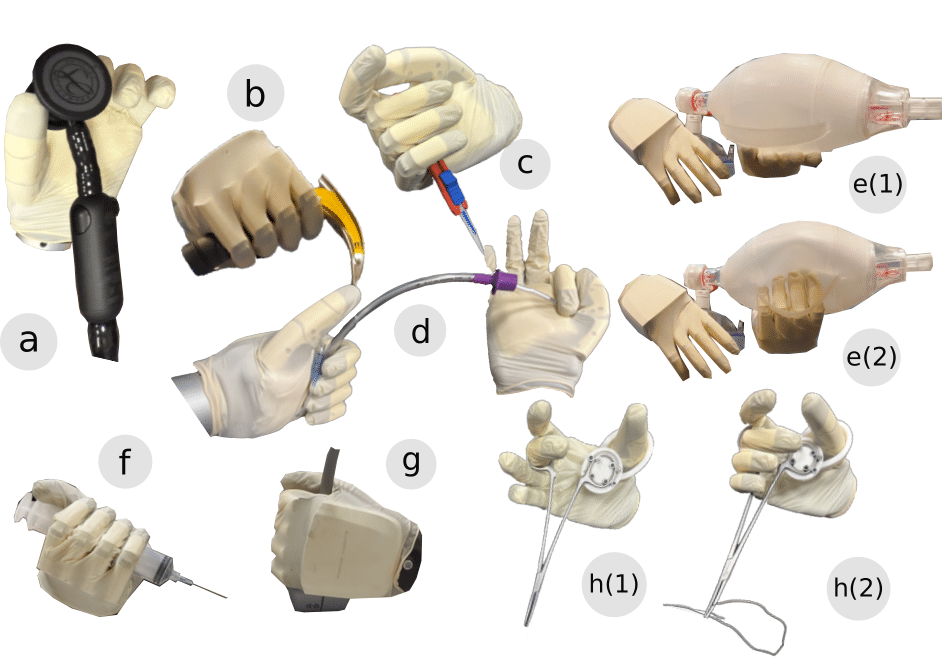} 
    \caption{\textbf{A visualization of grasping poses,} holding (a). Stethoscope, (b). Laryngoscope, (c). Scalpel, (d). The left hand is holding the Endotracheal tube, and the Right hand is holding the stylet (e(1)). Open hand holding the bag for bag-valve-mask (BVM) ventilation, (e(2)). Closed hand, squeezing their out of bag for BVM, (f). a needle syringe, (g). an ultrasound probe, h(1)-(2). opening and closing the surgical clamp to grasp a suture needle}
    \label{fig:grasp_poses}
    \vspace{-1em}
\end{figure}

A teleoperational system to control Unitree's G1 Humanoid Robot \cite{unitree2025g1} with Inspire Hands Gen4 \cite{inspire2025dexterous} was developed \autoref{fig:teleop_pipeline}.
Our teleoperation system is designed to control two arms simultaneously for contact-rich tasks, grasping and manipulating medical tools (see Fig.~\ref{fig:grasp_poses}). To handle force-sensitive procedures, such as positioning an ultrasound probe, an impedance controller is implemented that allows the operator to command tool position while automatically regulating the applied force to maintain safe and effective contact. In addition, this setup is augmented with a virtual spring-damper mechanism that couples the two arms, ensuring synchronized motion and force sharing during cooperative tasks (e.g., steadying a mask with one arm while using the other to actuate a pump). This layered control strategy provides flexible and compliant behavior, enabling robust bimanual coordination in complex medical scenarios.

\subsection{Operator Control Interface}

The operator control interface for the system requires high-fidelity bimanual input and an input for numerous grasping configurations.
Previous works have used VR headsets as an interface that provides an immersive visual experience \cite{cheng2024opentelevisionteleoperationimmersiveactive}.
However, hand controllers only support basic pose tracking and have a limited field of view; they cannot support the large catalog of grasping configurations.
Alternatively, consoles such as the da Vinci Research Kit \cite{kazanzides2014open} offer more precise pose inputs and foot pedals for convenient clutching, but are still limited in grasping configurations by the single pincher input for grasping.
To achieve robust and precise bimanual coordination with medical tools, a control interface integrates HTC Vive trackers \cite{viveProUserGuide}, multi-camera hand pose tracking, and a dual-pedal foot interface \autoref{fig:teleop_pipeline}. Each pedal is customizable and can either enable the clutching of one or both arms or activate a virtual spring-damper mode for tasks requiring bimanual engagement (e.g. pumping). In procedures such as suturing and tracheostomy, selectively clutching each arm facilitates fine-grained reorientation in constrained workspaces and mitigates excessive hand stress.


To refine the HTC Vive pose output and obtain multifinger grasping configurations of the operator, WiLoR~\cite{potamias2024wilor} was integrated, a state-of-the-art hand localization and reconstruction model. WiLoR operates within a multicamera setup, capturing the operator's hands from various perspectives.

Human hand keypoints are converted to robot joint angles by kinematic re-targetting \cite{qin2024anyteleopgeneralvisionbaseddexterous}, which optimizes hand-to-robot mapping for various gripper morphologies. It solves
\begin{equation}
\label{eqn:retarget}
\min_{q_t} 
\sum_{i=0}^{N} \Bigl\|\alpha\,v_{i}^{t} \;-\; f_{i}(q_{t})\Bigr\|^{2}
\;+\;
\beta \,\bigl\|q_{t} - q_{t-1}\bigr\|^{2},
\end{equation}
where $q_{t}$ are robot joint angles, $v_{i}^{t}$ the $i$-the human hand keypoint vector, $f_{i}(\cdot)$ maps joint angles to a corresponding end-effector vector~\cite{qin2024anyteleopgeneralvisionbaseddexterous, 9197124} and $\alpha$ is 1.5 for Inspire hands Gen4. 

To improve coverage of the operator's hands, two cameras, $\mathcal{C}_1$ and $\mathcal{C}_2$, are placed around the workspace. Each camera runs WiLoR independently \cite{potamias2024wilor} to detect keypoints using the MANO model \cite{MANO:SIGGRAPHASIA:2017}. For each camera, the viewing angle is measured $\phi_i$ between its optical axis and the estimated surface normal of the hand, yielding a cosine-based reliability factor:
\begin{equation}
\rho_i 
= 
\frac{\cos(\phi_i)}
{\cos(\phi_1) + \cos(\phi_2)}
\end{equation}
The fused keypoint $v_{\mathrm{fused}}$ is then calculated as,
\begin{equation}
    v_{\mathrm{fused}}
    =
    \rho_1\,\,v_1
    +
    \rho_2\,\,v_2
\end{equation}
Larger $\rho_i$ values give more weight to favorable camera views, producing robust 3D pose estimates under varied orientations or partial occlusions.






\subsection{Humanoid Teleoperation}

    

Our system applies \emph{relative} hand motions to drive the end-effector. When the operator presses a clutch, the current hand pose and the corresponding end-effector pose are saved. After releasing the clutch, subsequent orientation and position changes in the operator’s hand are mapped to the end-effector as follows:
\begin{equation}
{}^{W}\mathbf{R}_{EE}(t) 
\;=\; 
{}^{W}\mathbf{R}_{EE,\text{saved}}
\;\;
\Bigl(
{}^{W}\mathbf{R}_{H,\text{saved}}^\top \,
{}^{W}\mathbf{R}_{H}(t)
\Bigr)
\label{eqn:rot_eef}
\end{equation}
\begin{equation}
{}^{W}\mathbf{p}_{EE}(t) 
\;=\; 
{}^{W}\mathbf{p}_{EE,\text{saved}}
\;+\;
\Bigl(
{}^{W}\mathbf{p}_{H}(t)
-
{}^{W}\mathbf{p}_{H,\text{saved}}
\Bigr)
\label{eqn:pos_eef}
\end{equation}
where, ${}^{W}\mathbf{p}_{H}(t)$ and ${}^{W}\mathbf{R}_{H}(t)$ denote the current hand position and orientation in the world frame, respectively. This formulation ensures that only the \emph{incremental} motion from the saved pose is reflected on the end-effector, allowing comfortable reorientation without imposing significant or unintended movements on the robot. 

Many delicate tasks (e.g., ultrasound insertion and suturing) require tight regulation of interaction forces to avoid excessive contact forces and ensure precision. To achieve this, an impedance control strategy is developed, wherein the robot’s end effector position is adjusted in response to measured forces.
This approach trades off strict position tracking in favor of safe and compliant interactions.
First, estimate the Cartesian end-effector force, $\mathbf{F}_{\mathrm{EE}}$, from the measured joint torques, $\boldsymbol{\tau}$, using the humanoid's robotic arm Jacobian $\mathbf{J}$.
We then add this force to the gravity compensation terms, $\boldsymbol{\tau}_g$, to compute the final commanded torque, 
\begin{equation}
\boldsymbol{\tau}_{\mathrm{cmd}}
=
\boldsymbol{\tau}_g
+
\mathbf{J}^\top
\, \mathbf{F}_{\mathrm{EE}},
\label{eqn:impedence}
\end{equation}
Therefore, ensuring accurate force control while allowing the manipulator sufficient freedom to adapt its pose under external loads.
We maintain safe interactions and precise force application even in confined or sensitive environments by modulating the system’s impedance (i.e. effective stiffness, damping, and inertia).


To coordinate the motions of two arms--a ``follower'' (left arm) and a ``desired'' (right arm)---we incorporate a virtual spring-damper element into the impedance framework(\autoref{eqn:impedence}), helping distribute force requirements and prevent motor overload. Concretely, to update the left arm’s end-effector pose,
\(
{}^W\mathbf{x}_{EE,L}(t),
\)
by referencing the right arm’s \emph{desired} end-effector pose,
\(
{}^W\mathbf{x}_{EE,R,\text{des}}(t):
\)
\begin{equation}
    \begin{aligned}
        {}^W\mathbf{x}_{EE,L}(t) 
        &= {}^W\mathbf{x}_{EE,R,\text{des}}(t) \\
        &\quad+ \lambda 
        \Bigl(
        {}^W\mathbf{x}_{EE,L}(t) - {}^W\mathbf{x}_{EE,R,\text{des}}(t)
        \Bigr) \\
        &\quad+ \beta 
        \Bigl(
        {}^W \dot{\mathbf{x}}_{EE,L}(t) - {}^W \dot{\mathbf{x}}_{EE,R,\text{des}}(t)
        \Bigr)
    \end{aligned}
\end{equation}d
where, ${}^W \mathbf{x}_{EE,L}(t)$ = $\bigl\{{}^W\mathbf{p}_{EE,L}(t),\,{}^W\mathbf{R}_{EE,L}(t)\bigr\}$ and ${}^W \mathbf{x}_{EE,R,\text{des}}(t)$ = $\bigl\{{}^W\mathbf{p}_{EE,R,\text{des}}(t),\, {}^W\mathbf{R}_{EE,R,\text{des}}(t)\bigr\}$ denote the follower (left) and desired (right) arm poses, respectively as shown in \autoref{eqn:rot_eef} and \autoref{eqn:pos_eef}, and \(\lambda\) and \(\beta\) modulate the virtual ``spring'' and ``damper'' effects (where $\lambda$ is 3 and $\beta$ is 0.5). This construction promotes smoother, more compliant bimanual interactions, mitigating high motor stresses and delivering the appropriate end-effector force for contact-rich medical tasks.

\subsection{Grasping Medical Tools}

Since medical tasks require high-precision grasping, noise from a teleoperation interface can lead to inconsistent grasping or force inaccuracies.
To mitigate these issues, \emph{pre-configured} grasp poses tailored to each task are employed.
During teleoperation, the user’s joint space configuration is matched to these stored templates:
\begin{equation}
\mathbf{q}_{\mathrm{follower},t} 
= 
\arg\min_{\mathbf{q} \,\in\, \mathcal{Q}_{\text{preconf}}}
\bigl\lVert \mathbf{q}_{\mathrm{user},t} - \mathbf{q}\bigr\rVert,
\end{equation}
where $\mathbf{q}_{\mathrm{user}, t}$ is the hand joint configuration retargeted from the user’s keypoints $v_{user}$ as referred in \autoref{eqn:retarget} and $\mathcal{Q}_{\text{preconf}}$ represents the set of pre-configured grasp poses as shown in Fig. \ref{fig:grasp_poses}. This approach provides \emph{consistent}, \emph{accurate} gripper alignments, ensuring the appropriate force and posture for delicate surgical tasks.

\section{EXPERIMENTAL RESULTS}

This study represents the first attempt to integrate humanoid robotics into hospital clinical workflows.
We focus on 7 tasks that span physical examinations, emergency care, and precision needle tasks.
Our results demonstrate humanoids' potential role in modern healthcare and their current limitations. All experiments were done with non-clinician operators.


\subsection{Physical Examination}

\begin{figure}[t]
    \centering
    \includegraphics[width=\linewidth]{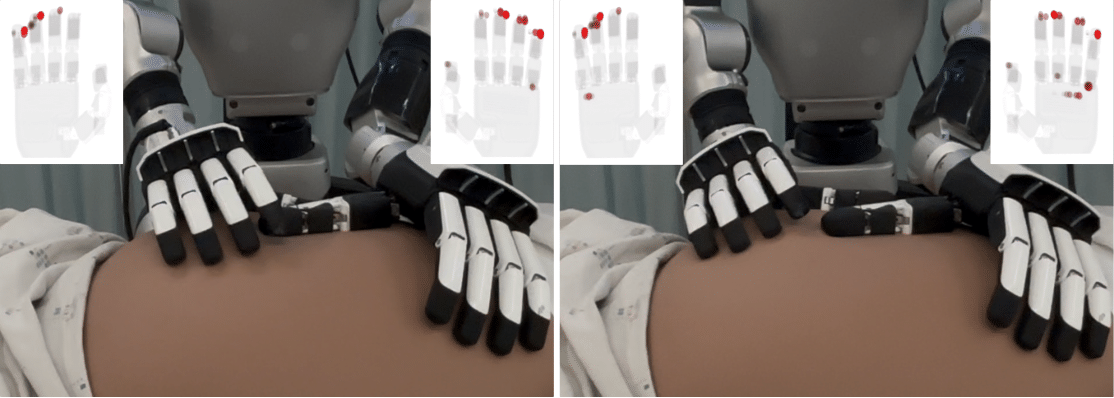} 
    \caption{\textbf{A close visualization of Leopold maneuver.} Robot pressing on pregnant simulator in the second maneuver position of Lepold Maneuvers with the pressure sensors that are activated depicted in the top corners of each image.}
    \label{fig:Lepold}
    \vspace{-1em}
\end{figure}

\textbf{Auscultation:} Auscultation, as shown in \autoref{fig:fig1}a, is the listening of sounds in the internal body such as the heart and lungs, and is conducted in almost all 500 million physical appointments per year \cite{CDC2023}.
In this study, the teleoperated humanoid robot performs auscultation using a digital stethoscope.
The examination is operated in two phases: first, the operator positions the stethoscope parallel to the patient's anatomical landmarks (e.g. chest and back) to establish proper orientation.
Second, the operator engages impedance control using the footpedal to regulate the position and applied force of the stethoscope on the patient's body, ensuring good contact.
This approach ensures consistent pressure and alignment, critical for capturing high-quality acoustic data, including frequency, intensity, duration, and quality of sounds. The robot was able to move through placements of the stethoscope along the Heart Auscultation path at about 6.35 seconds per realignment of the stethoscope. A realignment was only considered complete when it was placed correctly in terms of position and orientation.

During the experiment, the stethoscope picked up vibrations from the humanoid, which contributed to the masking of signal. This informs the potential need to characterize the vibrational characteristics of the robot actions would serve to mask out its effects of sensitive measurement tools such as the stethoscope, and further knowledge of the frequency domains of anatomical measurements such as heart-rate and breathing (both typical and atypical) would also provide additional separation of signal from noise.
\begin{figure}[t]
    \centering
    \begin{subfigure}[b]{0.48\textwidth}
        \centering
        \includegraphics[width=\textwidth]{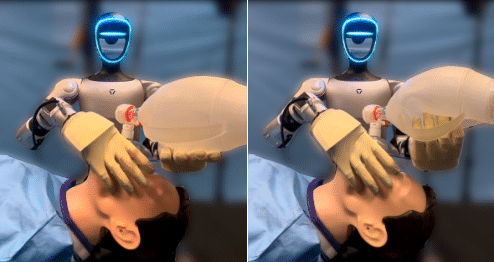} 
        \caption{A figure of the teleoperated humanoid system performing the BVM procedure. The left shows the single hand technique where the robot places the mask and holds it on the face of the patient. The right image shows the robot squeezing the bag to push air into the lungs of the patient}
        \label{fig:bvm}
        \vspace{0.3em}
    \end{subfigure}
    \hfill
    \begin{subfigure}[b]{0.48\textwidth}
        \centering
        \includegraphics[width=\textwidth]{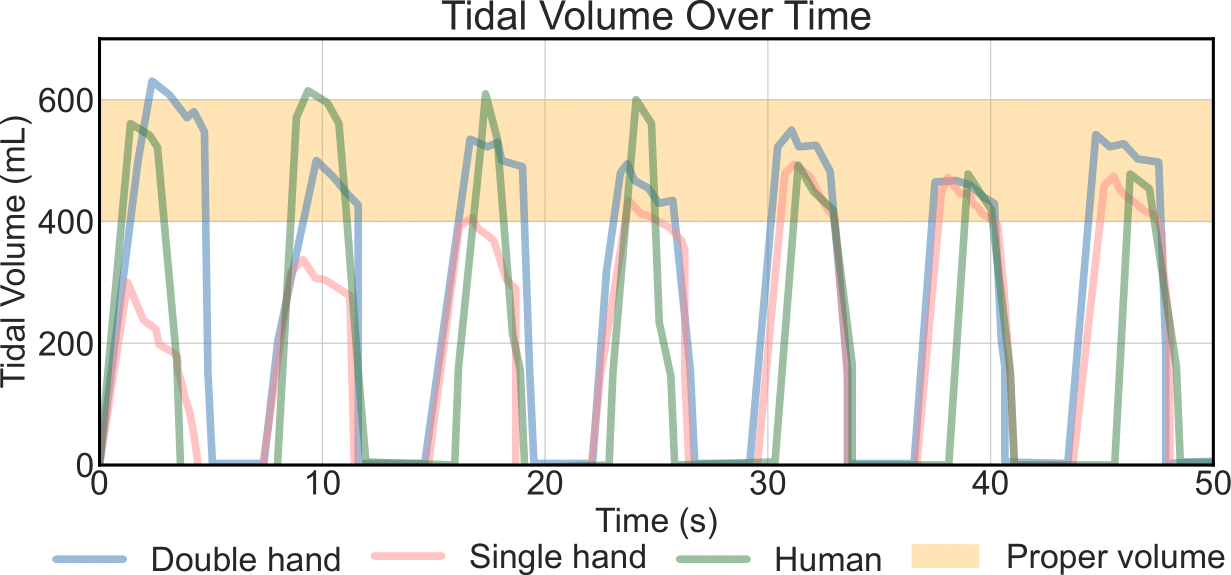}
        \caption{Quantitative tidal volume measurement of the system performing BVM. The system demonstrates better consistency than a human operator in ventilation timing.
        }
        \label{fig:bvm_quantitative}
    \end{subfigure}
    \caption{\textbf{Results of the BVM experiments.} The first row shows a qualitative visualization of the robot performing BVM independently. Second row evaluates BVM in 3 scenarios: robot alone, robot with human, and human alone.}
    \label{fig:ultrasound}
    \vspace{-2em}
\end{figure}

\begin{figure*}[h]
    \centering
    \begin{subfigure}[b]{0.9\textwidth}
        \centering
        \includegraphics[width=\textwidth]{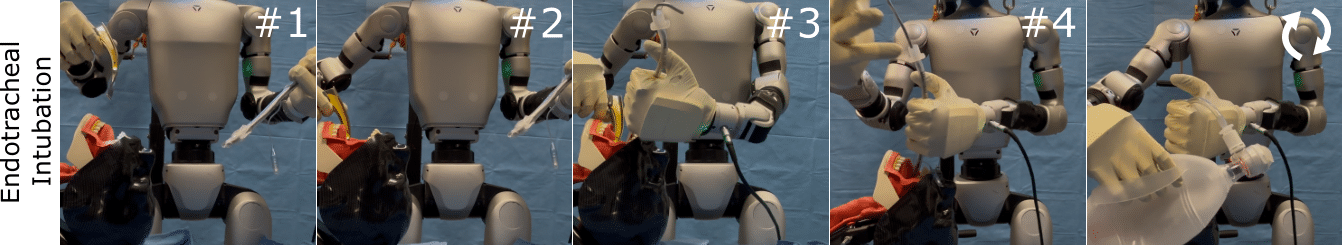} 
        \caption{A visualization of the endotracheal intubation approach. It includes (1). alignment of the laryngoscope to the mouth, (2) insertion of the laryngoscope to expose the upper trachea, (3). insertion of an endotracheal tube, (4). removal of the stylet, and then pumping of the air bag at the end.}
        \label{fig:endo_intubation}
    \end{subfigure}
    \hfill
    \begin{subfigure}[b]{0.9\textwidth}
        \centering
        \includegraphics[width=\textwidth]{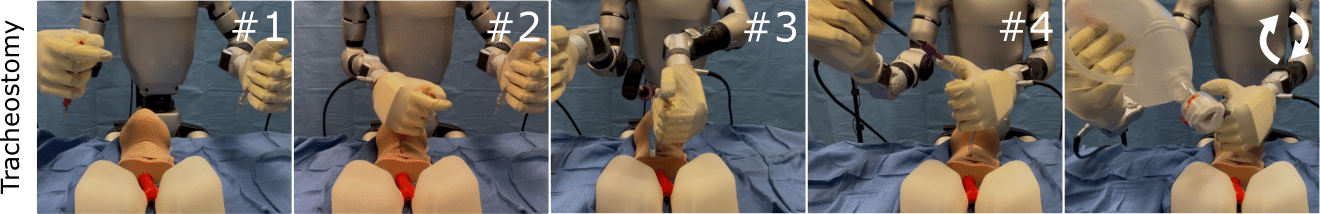} 
        \caption{A visualization of the tracheostomy approach. It includes (1). grasping of a scalpel and an endotracheal tube, (2) cutting to create an incision in the throat, (3). placement of the tube into the trachea, (4). removal of the stylet, and then pumping of the air bag at the end.}
        \label{fig:treacheostomy}
    \end{subfigure}
    \hfill
    \begin{subfigure}[b]{0.89\textwidth}
        \centering
        \includegraphics[width=\textwidth]{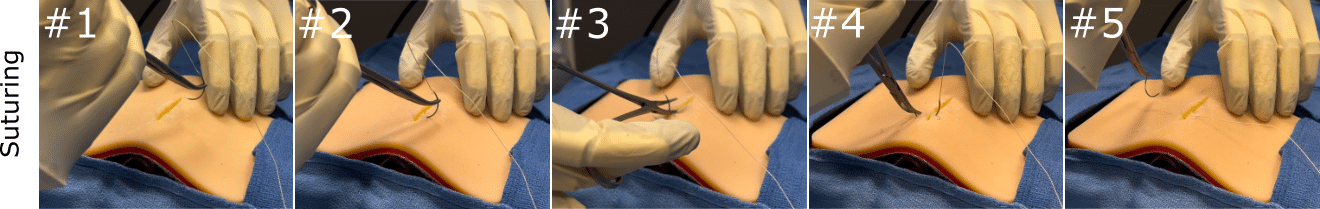} 
        \caption{A visualization of the system performing a suture throw. It includes (1). initial grasping of the suture needle, (2). insert the needle into the tissue and drive it, (3). releasing the needle, (4). re-grasping of the needle on the opposite side, and (5). pulling the thread through.}
        \label{fig:suturing}
    \end{subfigure}
    \caption{\textbf{A visualization of progression of 3 tasks.} From top to bottom is a time-series of the robot completing endotracheal intubation, tracheostomy, and suturing.}
    \label{fig:mainfig}
    \vspace{-2em}
\end{figure*}

\textbf{Leopold maneuvers:}
With approximately 3.5 million births occurring each year in the United States, where Leopold maneuvers are routinely performed as part of standard obstetrical examinations \cite{CDC2020Births}, the study explored the feasibility of replicating these maneuvers using a teleoperated humanoid system.
In the experiments, the humanoid was controlled to press around the baby using the four classic maneuvers: fundal grip, lateral grip, Pawlk grip, and pelvic grip to assess fetal lie, presentation, and engagement on a maternal simulator.
The system was able to successfully replicate the sequence of maneuvers.
However, the preliminary results indicate that further refinement in sensor sensitivity and control algorithms is necessary to match the nuanced palpation skills of experienced clinicians.
The geometry of the robot hands made it difficult to get pressure data from both the thumb and the other fingers, which led to the robot struggling to fully replicate what a clinician would be able to do.

\subsection{Emergency Interventions}

In emergency scenarios where a patient’s ability to breathe is compromised, effective airway management becomes critical.
Situations include respiratory distress, cardiopulmonary resuscitation, and when transitioning to advanced airway management \cite{lee2013performances}.
We explore a progression of interventions for airway management using a humanoid system.


\textbf{Bag Valve Mask Ventilation (BVM):}
BVM ventilation serves as the initial non-invasive method to provide oxygenation.
For example, 2.6\% of births use BVMs for resuscitation \cite{neonatal_resuscitation}.
BVM ventilation involves creating a tight seal between the mask and the patient’s face, then delivering oxygen by repeatedly squeezing a self-inflating bag.
In the first approach, \textit{single hand}, the humanoid uses one hand to manage the seal and the second hand to squeeze the bag.
\autoref{fig:bvm} shows an example of the humanoid pressing the mask on a manikin's mouth and nose, with another hand squeezing the air bag. While this is possible in real-life, it is difficult even for humans to perform solo. In clinical practice, it is recommended to perform BVM with 2 participants \cite{jesudian1985bag}, with the most experienced holding the mask against the face of the patient and the other pumping. Therefore, a second approach was developed, \textit{double hand}, where both humanoid hands squeeze the bag, with a human participant maintaining the seal.

Performance is quantified in three ways: ventilation interval (5-6 seconds desired for adults), ventilation time (around 1 second), and accuracy of tidal volume (400-600 mL per breath).
Quantitative data was measured with the Airway Management Simulator from BT Inc.
\autoref{fig:bvm_quantitative} shows the measurement of the tidal volume of a continuous 7 breaths (50 seconds) out of 15 breaths were recorded.
Our ventilation interval and time are 6s and 0.99s for using a single hand, 6s and 1.06s for using both hands.
The average tidal volume delivered is 471 and 533 mL for a single hand and both hands, with 86.7\% and 93.3\% of breaths within the desirable range, respectively.
A comparison of human performance from three participants was 4.8s, 4.9s, 4.9s for ventilation interval, 0.82s, 0.79s, 0.91s for ventilation time, and 93\%, 53\%, 80\% for tidal volume accuracy. The humanoid system thus, unsurprisingly, provides more consistent ventilation over humans.
While the approach was successful in bag squeezing for delivering oxygen, as expected, it was a challenge to maintain the seal between the bag and the patient in the single-hand approach.


\textbf{Endotracheal Intubation (EI):}
If the patient cannot sustain adequate breathing, EI becomes necessary to secure the airway with greater security and is reported to occur 13-20 million times per year \cite{Nadeem2017}.
In this procedure, a tube is inserted into the trachea through the mouth or nose to maintain a secure airway.
The tube was inserted through the mouth with a laryngoscope to expose the vocal cords and upper trachea as shown in \autoref{fig:endo_intubation}.
Once the tube is secured, the robot connects it to a mechanical ventilator and begins pumping at 10-12 pumps per minute with a tidal volume of between 400-600ml per pump, same as BVM.

Similar to BVM, the system successfully regulates the oxygen, however, a large amount of force is needed to open and hold the human mouth open.
Previous works have estimated the force required at 44N \cite{Wong2011}, which is significantly greater than the maximum force of the Unitree's G1 Humanoid Robot and thus these types of procedures require much stronger humanoids. Nonetheless, with the motion and orientation correct, some human assistance was provided in the direction of pulling on the laryngoscope to generate enough lifting force to open the airway. Subsequently, the robot was able to place the tube by itself in the trachea.
%

\textbf{Tracheostomy:} In cases where EI is unfeasible due to anatomical or pathological constraints, tracheostomy emerges as the final solution to establish a direct airway through a surgical opening in the trachea.
100,000 tracheotomies are performed every year \cite{Mehta2019}, and the first step is cutting an entry point through the neck to the trachea.
An endotracheal tube with a curved metal stylet is subsequently inserted into the trachea.
The metal stylet, which held rigidity to the tube, is removed and ventilation is applied to provide oxygen to the patient.
\autoref{fig:treacheostomy} shows a tracheostomy performed by the system. 
There are in total 3 different hand poses for both robot hands necessary for completing this procedure. They include poses of controlling a scalpel, holding the endotracheal tube, and removing the stylet.
Quantitative evaluations are conducted on the independent cutting and the tube insertion task. 

The cutting results are categorized as follows: 30\% successful, 40\% partially successful, and 30\% unsuccessful. The evaluation metric defines success as an incision exceeding 2 cm, partial success as an incision measuring less than 2 cm, and failure as any occurrence of unintended tissue damage. According to the NIH \cite{NRP2021} about 30 seconds recommended time for intubation specifically for preterm infants. The average time for our teleoperated humanoid to achieve a successful or partially successful intubation is 19 seconds.
Insertion is evaluated by the time it takes to insert the tip into an ideal opening. The average time is 18.4 seconds.  
The cutting and insertion results are based on a total of 10 trials.

Our tracheostomy approach was successful in creating a linear incision and inserting the tube tip promptly but revealed limitations for real-world application. Key challenges include insufficient force and tactile feedback for precise tissue tensioning and cutting, as well as the robot's limited dexterity and force to fully insert the tube into the narrow, high-friction tracheal simulator. These findings underscore the need for improved robotic force modulation, tactile feedback systems, and more realistic simulators to enhance performance and reliability in clinical settings.

\begin{figure}[t]
    \centering
    \includegraphics[width=\linewidth]{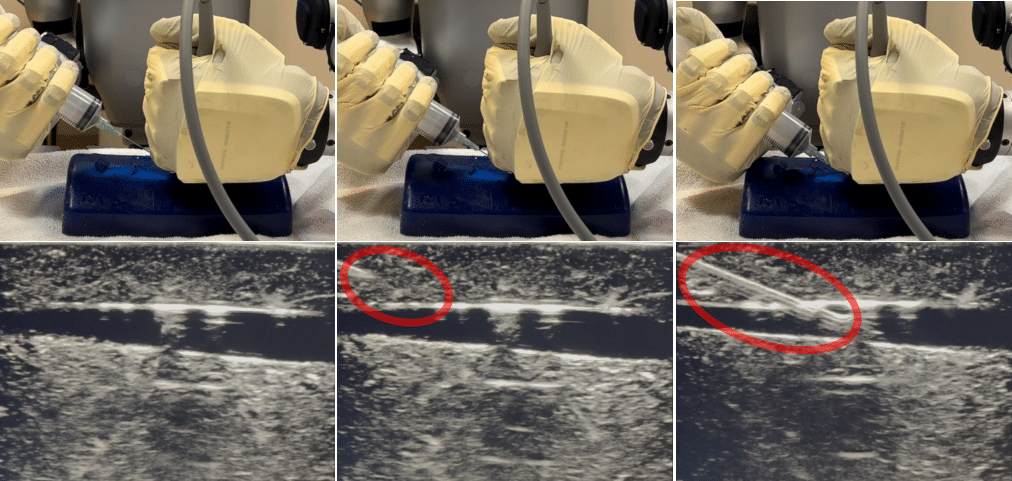} 
    \caption{\textbf{A visualization of ultrasound-guided injection.} The process of aligning the needle in the plane of the ultrasound, the initial entrance into the phantom, and finally entering the target location.}
    \label{fig:ultrasound_injection}
    \vspace{-1em}
\end{figure}
\subsection{Precision Needle Tasks}

\textbf{Ultrasound-guided Injection:} Over 1 million soft tissue filler injections are performed a year \cite{AestheticSociety2021}, an operation that frequently uses ultrasound for guidance.
This task is shown in \autoref{fig:ultrasound_injection} and involves bimanual coordination to hold an ultrasound probe for imaging and insert the needle into the target location.
The ultrasound probe first finds the anatomical structure of interest, in this case, a vessel, and aligns it with the image plane.
Under ultrasound guidance, the second hand advances the needle toward the target while using the ultrasound images for feedback.
\autoref{fig:ultrasound_injection} shows an example execution of the ultrasound-guided injection procedure. 
During needle insertions, the pose of the hand holding the syringe is regulated so that it is in-plane with the ultrasound and at a 30-degree angle to the top face of the tissue phantom surface.

We repeat this procedure 20 times with a non-clinician operator, and the results are shown in \autoref{tbl:ultrasound_injection}. 
An insertion is \textit{Hit} if the needle tip appears in the ultrasound image and stops inside the vessel structure, which is found to be 45\% in the experiment.
We report another 25\% \textit{Hit-Adj} when the needle tip either under or over-shoots, but can be re-adjusted to be within the vessel. 
We count 30\% \textit{miss} when no needle tip is observed on ultrasound. Counting both \textit{Hit, Hit-Adj}, there was 70\% success rate in total.
Our results are still less than the 90\% success rate of experienced clinicians reported by a previous work \cite{Hanada2017}. In \cite{Jagoda2023} where untrained unsupervised medical students attempted ultrasound-guided needle injection they had a success rate of 36.4\%.
The procedure in this paper was done by non-trained and non-medical personnel, and with more training, a higher success rate can be accomplished.

\begin{table}[t!]
\setlength\tabcolsep{.5em}
\centering
\caption{Overall success rate of 20 ultrasound-guided insertion experiments performed by the proposed humanoid teleoperation system.}
\begin{adjustbox}{width=0.45\textwidth}
    \begin{tabular}{ccc|c|cc}
    \toprule
 Miss & Hit-Adj & Hit & Success & Student\cite{Jagoda2023} & Doctor\cite{Hanada2017} \\
  \midrule
 30\% & 25\% & 45\% & 70\%& 36.4\% & 90\%\\
\bottomrule
\end{tabular}
\end{adjustbox}
\label{tbl:ultrasound_injection}
\vspace{1mm}
"Student": medical students who received no training. \\"Doctor": experienced anesthesiologists.
\vspace{-1em}
\end{table}
\begin{table}[t!]
\setlength\tabcolsep{0.8em}
\centering
\caption{Results of performing 16 suture throw experiments using the proposed teleoperation system. 
}
\begin{adjustbox}{width=0.45\textwidth}
    \begin{tabular}{c|ccc|cc}
    \toprule
  &  \multicolumn{3}{c|}{Stage success rate} & \multicolumn{2}{c}{Overall}\\
  \midrule  
  & Insert & Release & Extraction & Success & Time  \\
  \midrule
\cite{sen2016automating} & 98\% & 100\% & 90.2\% & 86.3\%& 112s\\
Ours & 50.0\% & 93.7\% & 100\%& 43.8\%& 141s\\
\bottomrule
\end{tabular}
\end{adjustbox}
\label{tbl:suturing_performance}
\vspace{-1em}
\end{table}

\textbf{Suturing:} Suturing is a common procedure for wound closures in trauma bays and surgery. 
This experiment demonstrates the capability of the proposed system in doing a suture throw on a training pad (\autoref{fig:suturing}).
In the demonstration, one hand presses the tissue phantom to keep it stable, and the other hand controls a surgical clamp that has a suture needle loaded.
The needle is ed and advanced through the tissue at a proper angle until the needle tip comes out on the opposite side.
Then the hand controls the clamp to release and re-grasp the needle tip to carefully pull out the suture needle, while the suture thread remains in the tissue.
To achieve releasing and re-grasping the suture needle, two robot hand poses are defined that control the opening and closing of the clamp.
The robot switches between them using the teleoperator's hand poses during the suturing process.

This procedure was repeated 16 times with a non-clinician, operator, and the results are shown in \autoref{tbl:suturing_performance}. It can be seen that there remains a large gap in performance between the humanoid and the specialized system with specific needle handling infrastructure. One contributing factor is that the proposed method had the highest failure rate during the insertion stage, mainly due to undesirable needle reorientation while navigating through tissue, when visibility is reduced. This suggests that haptic feedback is essential for precise needle tasks.



\vspace{-2mm}
\section{Discussion}

We conducted an exploratory study on the use of humanoid robots in hospital settings, leveraging the teleoperation system. This system directly mirrors the operator's hand motions, allowing the humanoid to grasp and manipulate medical tools. Additionally, it can engage an impedance control mode to regulate both position and applied force, enabling delicate interactions with patients and equipment.
Through this system, a wide range of medical tools that are commonly used were manipulated. Similar to applications in assembly tasks, the results demonstrate the potential for humanoid robots in clinical environments, as they can directly interface with the equipment used by medical staff.

For repetitive tasks such as bag-valve mask (BVM) squeezing, the system excels, as this motion can be directly programmed. Additionally, the system shows strong performance in aligning tools for ultrasound-guided injections, ensuring accurate positioning of the ultrasound probe and needle. Furthermore, the admittance control mechanism allows precise regulation of position and applied force, enabling controlled interactions with medical tools such as stethoscopes and ultrasound probes.
Despite these advancements, there remain key challenges in adapting the humanoid system for clinical use. One primary limitation is the system’s strength, as observed in the EI experiment, which restricts its ability to perform force-intensive tasks. Additionally, vibrations generated by the system can introduce artifacts, such as unwanted noise in stethoscope recordings. Finally, complex procedural tasks requiring intricate dexterity, such as Leopold’s maneuvers and suturing, remain difficult to execute effectively.

\section{CONCLUSION}

This exploratory study demonstrates the potential of humanoid systems to perform a variety of clinical tasks in hospital environments. The experiments, which evaluated procedures including auscultation, bag-valve mask ventilation, endotracheal intubation, tracheostomy, ultrasound-guided injection, and suturing, reveal that humanoid robots have a place in the future of medical procedures. In some areas, such as bag-valve mask ventilation, quantitative measurements already indicate that the system is capable of maintaining consistent ventilation intervals and tidal volumes, highlighting its promise in reducing operator variability and improving procedural accuracy in high-stakes scenarios.
Our findings also underscore important limitations. Mechanical constraints, such as insufficient force output for certain tasks, require human assistance in some procedures. The exploratory nature of this work and the reliance on teleoperation point to the need for further research to enhance autonomous functionality and integrate such systems seamlessly into clinical workflows.

Our study contributes to the growing body of evidence that supports the feasibility of humanoid robots in healthcare. By addressing both technical challenges and practical application scenarios, this work lays the foundations for future advancements aimed at mitigating workforce shortages and improving patient outcomes in hospital settings. Future research should focus on refining system capabilities, expanding the range of automated clinical tasks, and conducting rigorous clinical trials to validate the efficacy and safety of these emerging technologies.



\balance
\bibliographystyle{ieeetr}
\bibliography{root}

\newpage

\end{document}